\documentclass[journal]{IEEEtran}
\usepackage[T1]{fontenc}
\usepackage{graphicx}
\usepackage{times}
\usepackage{helvet}
\usepackage{courier}
\usepackage{amsmath}
\usepackage{algorithm}
\usepackage{algorithmic}
\usepackage{csquotes} 
\usepackage{color}
\usepackage{paralist}
\usepackage{amssymb}
\usepackage{indentfirst}
\usepackage{subfigure}
\usepackage{float}
\usepackage{multirow}
\usepackage{cite}
\usepackage{mathrsfs}

\usepackage{mathrsfs} 
\usepackage{amsfonts}
\usepackage{todonotes}
\usepackage{pgfplots} 
\usepackage{epstopdf}
\usepackage{epsfig}
\pgfplotsset{compat=newest}
\usepackage{color}
\definecolor{forestgreen}{RGB}{0,139,69}

\usepackage{xcolor}
\definecolor{citecolor}{HTML}{0071bc}
\usepackage[colorlinks, linkcolor=red,  anchorcolor=blue, citecolor=citecolor]{hyperref} 

\usepackage{xcolor}
\definecolor{SeaGreen4}{RGB}{0,205,102} 
\definecolor{SlateBlue}{RGB}{106,90,205} 
\definecolor{DarkRed}{RGB}{178,34,34} 
	
\usepackage{bbding}
\usepackage{url}

\usepackage{caption}
\usepackage{textcomp,booktabs}
\usepackage{amssymb}
\usepackage{pifont}

\usepackage{makecell}
\usepackage{colortbl}
\definecolor{mygray}{gray}{.9}
\definecolor{mypink}{rgb}{.99,.91,.95}
\definecolor{mycyan}{cmyk}{.3,0,0,0}

\begin{document}

\title{ SSTFormer: Bridging Spiking Neural Network and Memory Support Transformer for Frame-Event based Recognition }   

\author{Xiao Wang, \emph{Member, IEEE}, Yao Rong, Zongzhen Wu, Lin Zhu, \\ Bo Jiang*, Jin Tang, Yonghong Tian, \emph{Fellow, IEEE} 

\thanks{$\bullet$ Xiao Wang, Yao Rong, Zongzhen Wu, Bo Jiang, and Jin Tang are with the School of Computer Science and Technology, Anhui University, Institute of Artificial Intelligence, Hefei Comprehensive National Science Center, Hefei 230601, China. (email: \{xiaowang, tangjin\}@ahu.edu.cn, \{rytaptap, 17398389386, zeyiabc\}@163.com)} 

\thanks{$\bullet$ Lin Zhu is with Beijing Institute Of Technology, Beijing, China (email: linzhu@pku.edu.cn)}

\thanks{$\bullet$ Yonghong Tian is with Peng Cheng Laboratory, Shenzhen, China, and National Engineering Laboratory for Video Technology, School of Electronics Engineering and Computer Science, Peking University, Beijing, China. (email: yhtian@pku.edu.cn)} 

\thanks{* Corresponding Author: Bo Jiang} 
} 



\maketitle

\begin{abstract}
Event camera-based pattern recognition is a newly arising research topic in recent years. Current researchers usually transform the event streams into images, graphs, or voxels, and adopt deep neural networks for event-based classification. Although good performance can be achieved on simple event recognition datasets, however, their results may be still limited due to the following two issues. Firstly, they adopt spatial sparse event streams for recognition only, which may fail to capture the color and detailed texture information well. Secondly, they adopt either Spiking Neural Networks (SNN) for energy-efficient recognition with suboptimal results, or Artificial Neural Networks (ANN) for energy-intensive, high-performance recognition. However, \textcolor{black}{few} of them consider achieving a balance between these two aspects. In this paper, we formally propose to recognize patterns by fusing RGB frames and event streams simultaneously and propose a new RGB frame-event recognition framework to address the aforementioned issues. The proposed method contains four main modules, i.e., memory support Transformer network for RGB frame encoding, spiking neural network for raw event stream encoding, multi-modal bottleneck fusion module for RGB-Event feature aggregation, and prediction head. Due to the \textcolor{black}{scarcity} of RGB-Event based classification dataset, we also propose a large-scale PokerEvent dataset which contains 114 classes, and 27102 frame-event pairs recorded using a DVS346 event camera. Extensive experiments on two RGB-Event based classification datasets fully validated the effectiveness of our proposed framework. We hope this work will boost the development of pattern recognition by fusing RGB frames and event streams. Both our dataset and source code of this work will be released at \url{https://github.com/Event-AHU/SSTFormer}. 
\end{abstract}


\IEEEpeerreviewmaketitle

\section{Introduction}

\IEEEPARstart{T}{he} mainstream video-based classification algorithms are widely developed based on RGB cameras. With the help of deep learning, many representative deep models are proposed, such as TSM~\cite{lin2019tsm}, C3D~\cite{tran2015c3d},   SlowFast~\cite{feichtenhofer2019slowfast}, DevNet~\cite{gan2015devnet}, TSN~\cite{wang2016temporal}, Attnclusters~\cite{long2018attention} and Video-SwinTrans~\cite{liu2021videoSwin}, etc. These works learn the deep feature representation well and contribute significantly to many practical applications and video-related tasks. However, these models may perform poorly in some extremely challenging scenarios (for example, fast motion, low illumination, and over-exposure) due to the utilization of RGB cameras. This is because RGB cameras adopt a global synchronous exposure mechanism and have a limited frame rate.

\begin{figure}
\center
\includegraphics[width=3.5in]{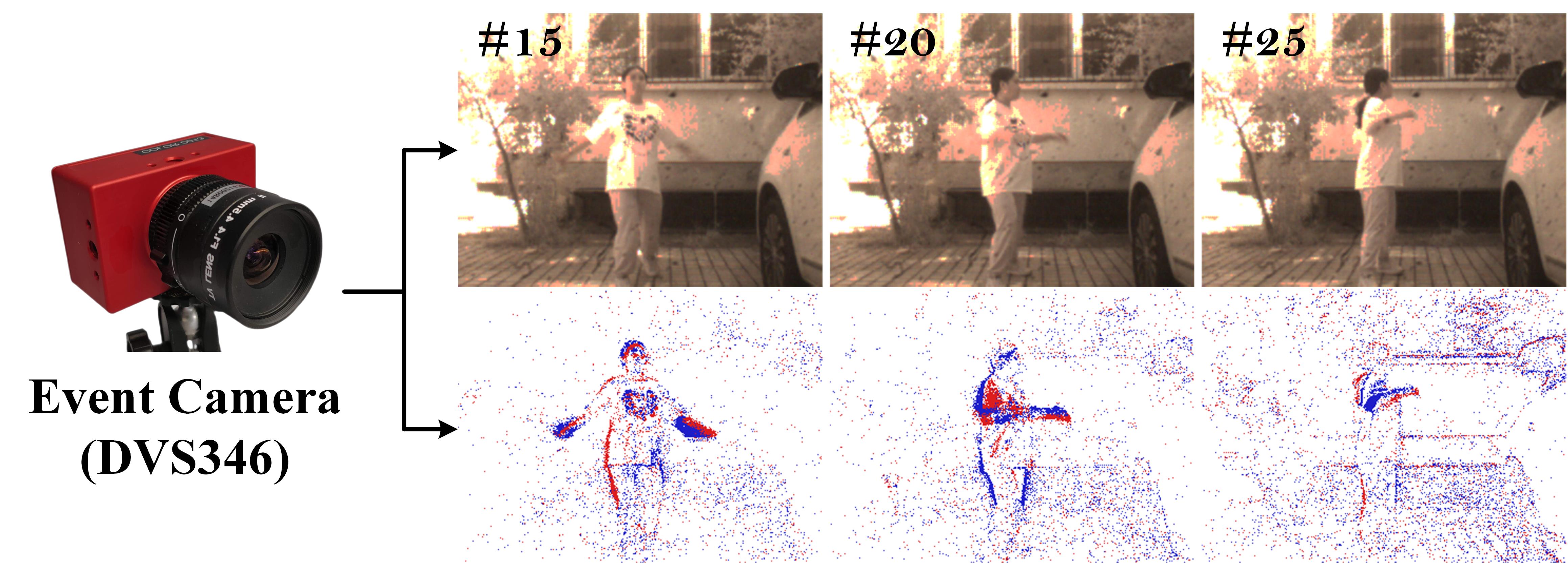}
\caption{Illustration of video classification by fusing RGB frames and event stream. 
            The event images (bottom row) are stacked for visualization.}   
\label{firstIMG}
\end{figure}

Recently, biologically inspired event cameras (also called DVS, Dynamic Vision Sensor) have attracted great interest~\cite{eTLD2021, EDFLOW2022} due to their unique imaging principle and key characteristics including \emph{high dynamic range, high temporal resolution, low latency, low power,} etc. Different from RGB cameras which record the light intensity of a scene in a synchronous way, each pixel in event cameras outputs an event spike only when the variation of light intensity exceeds a certain threshold. The event that corresponds to a decrease in brightness is termed \emph{ON} event, otherwise, \emph{OFF} event. Usually, each event can be represented as a quadruple \{$x, y, t, p$\}, where $x, y$ \textcolor{black}{denote} the spatial coordinates, $t$ is the timestamp, and $p$ is the polarity (1 and -1 are used to denote the \emph{ON} and \emph{OFF} event, respectively). Event cameras have been applied in many computer vision tasks, such as object detection, visual tracking~\cite{wang2021visevent, tang2022coesot}, pose estimation, etc. 
\textcolor{black}{The RGB and Event scenes synchronously recorded by the DVS346 camera are shown in  Fig.~\ref{firstIMG}.}

There are some works, such as Graph Neural Networks (GNN)~\cite{xie2022vmvgcn, schaefer2022aegnn}, Convolutional Neural Networks (CNN)~\cite{wang2019evGait, xu2015discriminative}, and Spiking Neural Networks (SNN)~\cite{zhou2023spikformer, fang2021PLIF} 
which have been proposed for pattern recognition of event cameras. 
For example, Asynchronous Event-based Graph Neural Network (short for AEGNN~\cite{schaefer2022aegnn}) is proposed to process events as spatio-temporal graphs in an incremental learning manner. Zhou et al. combine the spiking neurons with Transformer network and propose the Spikformer~\cite{zhou2023spikformer} for event data representation and  recognition. Wang et al. propose a CNN-based model for gait recognition, termed EV-Gait~\cite{wang2019evGait}. 
However, we believe these works may be limited due to the following issues. 
\textbf{Firstly}, existing works focus on designing pure SNN or Artificial Neural Networks (ANN) for event-based recognition. The SNN performs better in saving energy consumption, while the ANN achieves a higher classification performance. However, the trade-off between these two kinds of neural networks is seldom considered. 
\textbf{Secondly}, these works are developed based on a single event camera only, meanwhile, the event stream is sparse in the spatial view. It also fails to capture the color and detailed texture information which are important for pattern recognition. As noted in many works~\cite{tang2022coesot, wang2021visevent, jiang2020cmsalgan, zhang2020multimodal}, it will be a better choice to combine multi-modal cues to obtain higher performance. 

To address the aforementioned two issues, in this paper, we propose a novel framework that integrates both visual frames and event streams for pattern recognition. Generally speaking, it mainly contains four main modules, i.e., Spiking Convolutional Neural Network (\textcolor{black}{SCNN}), Memory Support Transformer Network (MST), Multi-modal Bottlenecks Fusion (MBF) module, and prediction head.  
Different from existing works that transform the event streams into an image-like representation~\cite{zhang2021fe108}, graph~\cite{schaefer2022aegnn}, or voxel~\cite{li2021graphevent}, we propose a Spiking Convolutional Neural Network (\textcolor{black}{SCNN}) that takes the raw event streams as input directly for energy-efficient perception. 
Also, multi-scale features are accumulated and enhanced by using an ANN decoder, therefore, our SCNN branch can achieve a better trade-off between the recognition performance and energy consumption. 
For the RGB frames, a novel Memory Support Transformer Network (MST) is proposed to encode the spatial-temporal information. After embedding the raw frames into feature maps, we first divide the feature maps into multiple clips in chronological order. 
\textcolor{black}{
Inspired by the few-shot learning~\cite{song2023FSLsurvey} which \textcolor{black}{boosts} the feature learning and prediction of query samples based on support samples, in this work, we treat the last feature in each clip as the query vector, and the others are combined with the memory propagated from the previous clip as the memory support feature. 
}
The Gated Recurrent Unit (GRU) network is used to capture the temporal information. Then, its output and the query vector are fed into the cross-attention module for adaptive fusion. 
Finally, a Multi-modal Bottlenecks Fusion (MBF) module is introduced to fuse the RGB and event features for pattern recognition. 
An overview of our proposed RGB-Event based recognition framework is shown in Fig.~\ref{framework}.

Moreover, considering that there is still a lack of large-scale RGB-Event classification dataset to support current experimental validation, this paper proposes a new dataset to bridge this data gap, termed \emph{PokerEvent}. It is recorded using a DVS346 event camera which outputs aligned RGB frames and event streams simultaneously. It contains 114 classes, and 27102 frame-event pairs and the resolution is $346 \times 260$. We split them into training, validation, and testing \textcolor{black}{subsets} which contains 16216, 2687, and 8199 samples, respectively.

Overall, the main contributions of this paper can be summarized as the following three aspects: 

$\bullet$ We propose a novel frame-event based pattern recognition framework, which mainly contains Spiking Convolutional Neural Network (\textcolor{black}{SCNN}), Memory Support Transformer Network (MST), and Multi-modal Bottlenecks Fusion (MBF) module. To the best of our knowledge, it is the first work to combine the RGB frame and event streams representations for classification, which achieves a better trade-off between energy consumption and recognition accuracy. 

$\bullet$ We propose a new large-scale RGB-Event classification dataset \emph{PokerEvent}, which contains 16216, 2687, and 8199 samples for training, validation, and testing. We believe that this dataset will greatly facilitate event camera related pattern recognition tasks. 

$\bullet$ Extensive experiments on two large-scale RGB-Event classification datasets fully validated the effectiveness of our proposed learning framework.

The rest of this paper is organized as follows: We give an introduction to related works in Section~\ref{RelatedWork}. The method we proposed in this paper is mainly described in Section~\ref{Approach}, with a focus on input representation, network architectures, and loss function. The experiments are conducted in Section~\ref{Experiments}. We describe the details of the newly proposed \emph{PokerEvent} dataset in Section~\ref{datasetPoker}. Finally, we conclude this paper in Section~\ref{Conclusion}.

\section{Related Work}\label{RelatedWork}

In this section, we give a brief review of Event-based Classification, Spiking Neural Networks, and Transformer Networks. 
More related works can be found in the following surveys\footnote{\url{https://github.com/Event-AHU/Event_Camera_in_Top_Conference}}~\cite{kong2018humanARSurvey, gallegoevent, ahmad2021graph}.

\subsection{Event-based Classification} 
The mainstream of event camera-based recognition can be divided into the following categories, i.e., CNN-based~\cite{wang2019evGait}, GNN-based~\cite{xie2022vmvgcn, schaefer2022aegnn}, and SNN-based~\cite{zhou2023spikformer, fang2021PLIF}. To be specific, 
Arnon et al.~\cite{amir2017low} propose the first gesture recognition system based on TrueNorth neurosynaptic processor. 
Xavier et al.~\cite{clady2017motion} propose an event-based luminance-free feature for local corner detection and global gesture recognition. 
Chen et al.~\cite{chen2021novel} propose a hand gesture recognition system based on DVS and also design a wearable glove with a high-frequency active LED marker that fully exploits its properties. 
A retino morphic event-driven representation (EDR) is proposed by Chen et al.~\cite{chen2019fast}, which can realize three important functions of the biological retina, i.e., the logarithmic transformation, ON/OFF pathways, and integration of multiple timescales. 
The authors of~\cite{lagorce2016hots} represent the recent temporal activity within a local spatial neighborhood and utilize the rich temporal information provided by events to create contexts in the form of time surfaces, termed HOTS, for the recognition task. 
Wu et al. first transform the event flow into images, then, predict and combine the human pose with event images for HAR~\cite{wu2020multipath}. 
Wang et al.~\cite{wang2019evGait} propose a CNN-based model for event-based gait recognition. 
Deng et al.~\cite{deng2021mvfnet} propose a multi-view attention-aware network that projects an event stream into multi-view 2D maps to make full use of existing 2D models and spatiotemporal complements. 
Liu et al.~\cite{Liu2024VMSTNet} propose the VMST-Net which is a voxel-based multi-scale Transformer network to process event streams.

Graph neural networks (GNN) and SNNs are also fully exploited for event-based recognition~\cite{george2020reservoir, bi2019graph, mehr2019action, samadzadeh2020convsnn, li2018deepCNN, ceolini2020hand, panda2018learning, liu2020unsupervised, xing2020new, chen2020dyGCN, wang2021eventGNN}. 
Specifically, Chen et al.~\cite{chen2020dyGCN} treat the event flows as 3D point cloud and use dynamic GNNs to learn the spatio-temporal features for gesture recognition. 
Wang et al.~\cite{wang2021eventGNN} develop a graph neural network for event-based gait recognition. 
Xing et al. design a spiking convolutional recurrent neural network (SCRNN) architecture for event-based sequences~\cite{xing2020new}.  
From these works, we can find that previous researchers adopt SNN for energy-efficient computation and satisfy the final results, or ANN for energy-intensive high-performance recognition. Different from these works, our proposed RGB-Event based recognition framework achieves a better tradeoff between energy consumption and recognition performance.

\subsection{Spiking Neural Networks} 
Due to the advantages of energy consumption, more and more researchers are devoted to the study of spiking neural networks~\cite{lobo2020SNNreview, wang2020SNNsurvey, zhou2021SNNIIR}. To be specific, 
Lee et al.~\cite{lee2020spikeflownet} jointly utilize the SNN and ANN for hierarchical optical flow estimation. 
Federico et al.~\cite{paredes2019SNNopticalflow} propose a hierarchical spiking architecture for optical flow estimation \textcolor{black}{with} selectivity to the local-global motion via SpikeTiming-Dependent Plasticity (STDP)~\cite{caporale2008STDP}. 
Zhou et al.~\cite{zhou2021SNNIIR} exploit the non-leaky IF neuron with single-spike temporal coding to train deep SNNs. 
Zhou et al. combine the spiking neurons with the Transformer network and propose the Spikformer~\cite{zhou2023spikformer} for recognition. 
Fang et al.~\cite{fang2021PLIF} propose a new method to learn both the synaptic weights and the membrane time constants of SNNs. 
\textcolor{black}{For multimodal fusion in spiking neural networks, researchers conduct some explorations. This paper~\cite{Liu2022Event-Based} presents an event-based end-to-end multimodal spiking neural network that utilizes the attention mechanism to achieve effective fusion of visual and auditory information. CMCI~\cite{Jiang2024CMCI:} is a robust multimodal fusion method based on spiking neural networks, and it conducts a systematic comparison on visual, auditory, and olfactory fusion recognition tasks.Besides, researchers also explore spiking neural networks based on events. Zhang et al.~\cite{zhang2022spiking} propose a tracking network based on SNN, which can effectively extract spatio-temporal information from events. Kugele et al.~\cite{Kugele2021Hybrid} propose a hybrid SNN-ANN model for event-based pattern recognition and object detection. EAS-SNN~\cite{Wang2024EAS-SNN:} is an end-to-end adaptive sampling and representation method for event-based detection. Aydin~\cite{Aydin2024A} propose an event-based hybrid ANN-SNN architecture, which achieves low-power consumption and low-latency visual perception.}  In addition to the standard recognition task, the SNNs are also widely used in many other tasks, such as visual tracking and scene reconstruction. Different from most of these works, in this paper, we propose a novel bottleneck fusion based SCNN-ANN framework for RGB-Event based recognition.

\subsection{Transformer Networks}
The Transformer proposed in the field of natural language processing achieved excellent performance~\cite{vaswani2017attention} and is also adapted in other research communities, like computer vision. To be specific, Vision Transformer (ViT)~\cite{dosovitskiy2020image} partitions the image into a series of non-overlapping patches and projects them into tokens as the input of self-attention modules. Their success is owed to the long-term relationship mining between various tokens. DeiT~\cite{touvron2021training} proposes a strategy based on token distillation using convolutional networks as the teacher network, which addresses the issue of requiring a large amount of data for pre-training. 
Chen et al.~\cite{Chen2023ECSNet} propose the 2D-1T event cloud sequence (2D-1T ECS) which is a compact event representation, and build a light-weight spatio-temporal learning framework (ECSNet) for event-based object classification and action recognition tasks. 
ConViT~\cite{d2021convit} introduces the inductive bias of CNN into the Transformer and achieves better results. VOLO~\cite{yuan2022volo} uses a two-stage structure, which can generate attention weights corresponding to its surrounding tokens through simple linear transformations, avoiding the expensive computational cost of the original self-attention mechanism. Swin Transformer~\cite{liu2021swin} introduces a sliding window mechanism, enabling the model to handle super-resolution images, solving the problem of high computational complexity in ViT model. Chen et al. propose the MoCo v3~\cite{chen2021empirical} by applying contrastive learning for ViT-based self-supervised learning. He et al. randomly mask a high-proportion of patches under the auto-encoder framework and propose the MAE~\cite{he2022masked} for self-supervised learning. Pre-trained multimodal big models like the CLIP~\cite{radford2021learning} are also developed based on vision Transformer backbone networks. More tasks are boosted by following the Transformer based feature learning framework, including image recognition~\cite{liu2021swin}, action recognition~\cite{bertasius2021space}, object detection~\cite{carion2020end}, semantic segmentation~\cite{cheng2021per}, etc. Different from these works, we propose a novel memory support Transformer that captures the spatiotemporal information for recognition.

\begin{figure*}
\center
\includegraphics[width=6.8in]{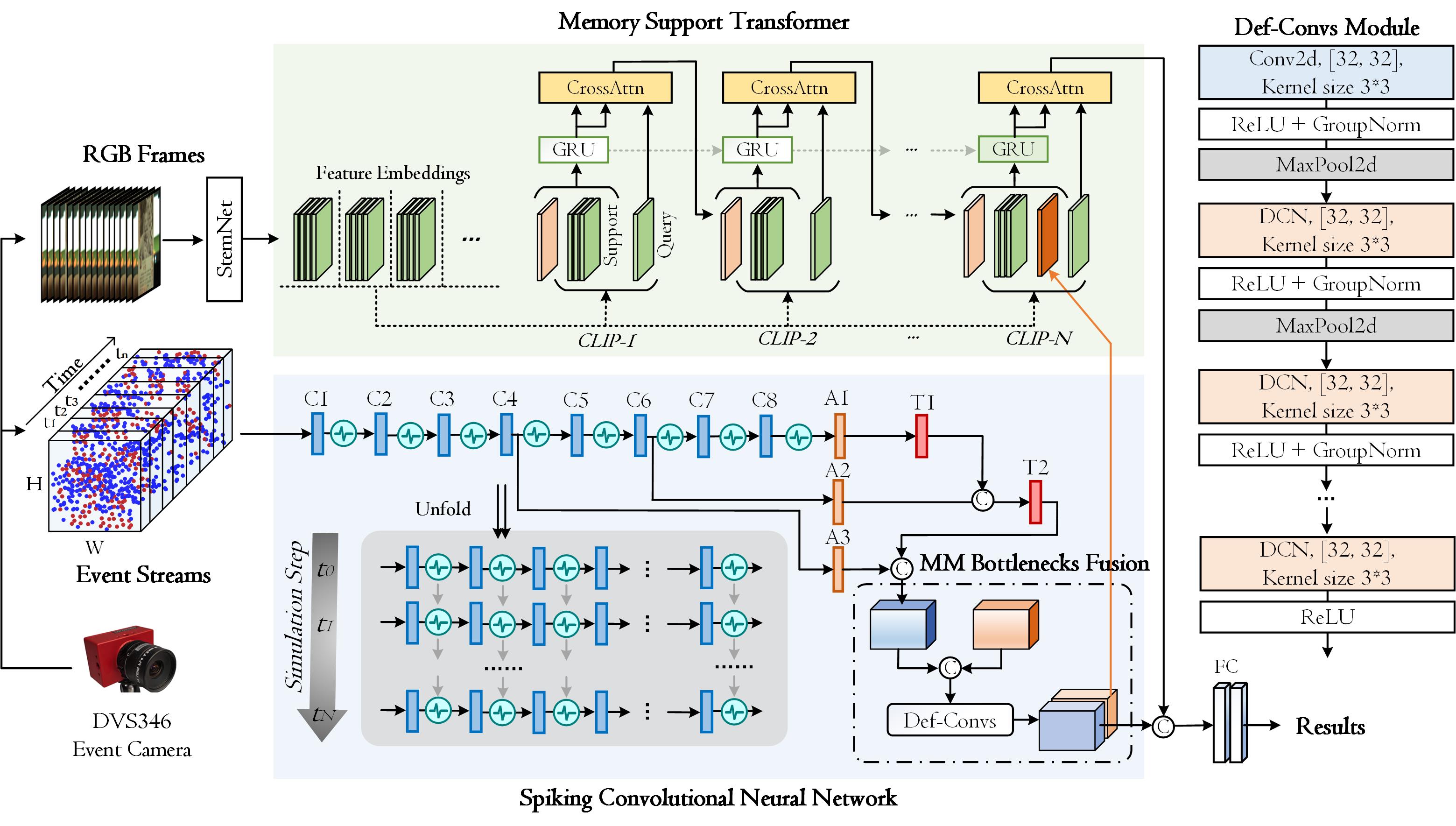}
\caption{{An overview of our proposed multi-modal fusion framework for RGB-Event based pattern recognition.} We propose a novel SCNN (Spiking Convolutional Neural Network) which is a hybrid SNN-ANN network to directly encode the raw event streams instead of pre-processing them into the expression of intermediate form. It achieves a better trade-off between the energy-consumption and overall recognition results. For the RGB input, we design a novel memory support Transformer (MST) network to learn the spatial-temporal information. We first divide the video frames into multiple clips, then, we treat the last frame in a clip as the query and the rest as support frames, and conduct support-query interactive learning using the cross-attention mechanism. The output of SCNN and MST modules are fused with bottleneck feature maps which are then fed into the prediction head for recognition.} 
\label{framework}
\end{figure*}

\section{Methodology} \label{Approach}  
In this section, we will first give an overview of our proposed RGB-Event recognition framework. Then, we will briefly introduce the representations of input RGB frames and event streams. Later, we will dive into the details of our method, with a focus on the Memory Support Transformer Network (MST), Spiking Convolutional Neural Network (\textcolor{black}{SCNN}), Multi-modal Bottlenecks Fusion (MBF) module, and prediction head. Then, we introduce the loss function used in the training phase. Finally, we introduce the dual-Transformer by integrating SpikingFormer and MST for RGB-Event based recognition.

\subsection{Overview} \label{SNNCNNoverview} 
\textcolor{black}{
As shown in Fig.~\ref{framework}, our proposed RGB-Event object recognition framework takes the RGB frames and event streams as the input. For the RGB frames, we first obtain their feature embeddings using a StemNet (ResNet18/ResNet50~\cite{he2016resnet} is used in our implementation) and propose to learn the spatial-temporal representations by adopting a support-query feature learning architecture. More in detail, for the divided frame clips, we first use GRU networks to encode the feature embeddings and previous memory output. Then, the cross-attention is employed to capture the dependencies between support and query features. 
For the event streams, a hybrid SNN-ANN network is proposed to directly process the raw event points. The \textcolor{black}{SCNN} encoder contains several spiking convolutional layers, meanwhile, the ANN decoder is proposed to further enhance the feature representation learning. Multimodal bottleneck features are proposed to enhance the interactive learning between the RGB frames and event streams. Finally, the output features are fed into two fully connected layers for RGB-Event based pattern recognition. 
In addition, we also exploit dual-Transformer based network architecture for RGB-Event based recognition by integrating SpikingFormer~\cite{zhou2023spikingformer} and our newly proposed MST module. 
}

\subsection{Input Representation} \label{inputRepres} 
Different from existing works that transform the event streams into an image-like representation~\cite{zhang2021fe108}, graph~\cite{schaefer2022aegnn}, or voxel~\cite{li2021graphevent}, we propose a Spiking Convolutional Neural Network (\textcolor{black}{SCNN}) that takes the raw event streams as input directly for energy-efficient perception. Given the event streams $\mathcal{E} = \{e_1, e_2, ..., e_T\}$, where $e_i$ denotes the event point and $T$ is the maximum number of event points. Usually, we use  quadruple $(x, y, t, p)$  to denote each event point, where $x, y$ are spatial coordinates, $t$ is the timestamp, and $p$ is the polarity. Due to the dense temporal information, in the scenes of long and fast motion, a very large number of events are captured. Therefore, previous researchers usually first down-sample the dense event streams into sparse ones using farthest point sampling technique~\cite{eldar1997fps}, or stack them into image-like representations~\cite{wang2019evGait}, voxels~\cite{li2021graphevent}, etc. We believe these operations may lose the temporal information to some extent. In this paper, we directly take the raw event points as the input and propose the Spiking Convolutional Neural Network (\textcolor{black}{SCNN}) to learn the spatial-temporal information. 
Below, we denote  $\mathcal{F} = \{f_1, f_2, ..., f_N\}$ as a visual video sequence with $N$ frames, where $f_i$ denotes the i$^{th}$ video frame.

\subsection{Network Architecture} \label{networkArch} 

As shown in Fig.~\ref{framework}, there are four main modules in our proposed framework, including Spiking Convolutional Neural Network (\textcolor{black}{SCNN}), Memory Support Transformer (MST), Multi-modal Bottleneck Fusion (MBF), and Prediction Head. We will introduce these modules in the following paragraphs respectively.

\begin{figure}[!htp]
\center
\includegraphics[width=2.8in]{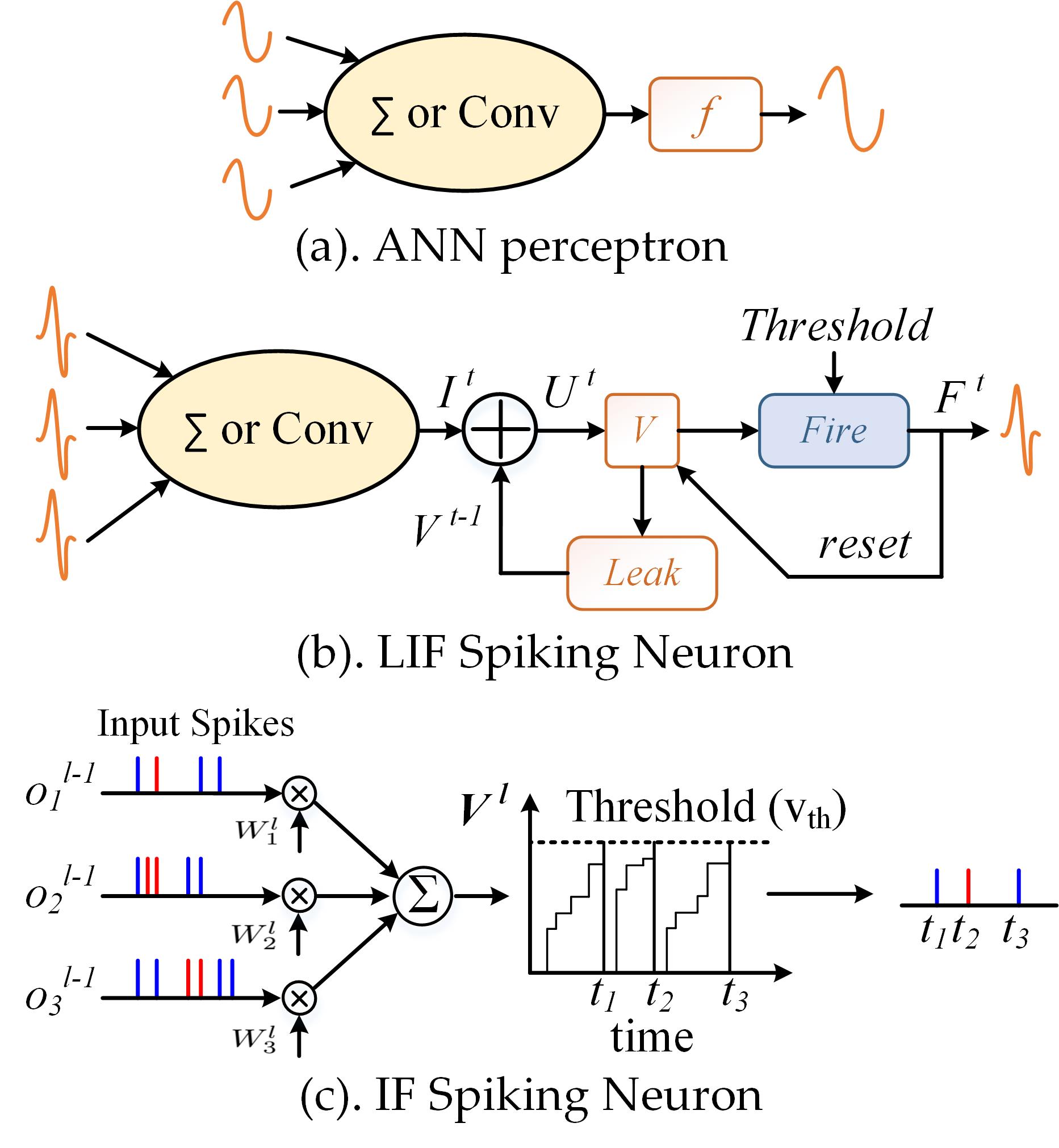}
\caption{Comparison between the widely used ANN perceptron, LIF, and IF spiking neuron. } 
\label{spikingNeuron}
\end{figure}

\noindent 
\textbf{Spiking Neural Network. } 
Given the event streams $\mathcal{E}$, we propose a novel Spiking Convolutional Neural Network (\textcolor{black}{SCNN}) to directly encode the raw event. Therefore, we do not need to perform special processing on the pulse signal, but can directly use the spatio-temporal modeling ability of SCNN for feature learning. Generally speaking, it contains two main parts, i.e., Spiking Convolutional Neural Network (SCNN) encoder and ANN decoder. More in detail,  SCNN encoder contains eight convolutional layers and a spiking neuron is used between every two layers. In the practical implementation, a variety of spiking neurons can be selected, such as the IF~\cite{cao2015spiking, diehl2015snnbalancing}, LIF~\cite{gerstner2002spiking}, and LIAF~\cite{wu2021liaf} neurons. 
For example, 
in LIF~\cite{gerstner2002spiking}, it computes the procedure of spiking neuron as: 
\begin{equation}
\label{lifneuron} 
\tau \frac{dU}{dt} = -U + RI, 
\end{equation}
where $I$ and $R$ denote the current and input resistance, respectively. $\tau$ is a time constant of the circuit. 
\textcolor{black}{$t$ and $U$ denote time and membrane potential.} 
\textcolor{black}{The aforementioned equation can be formulated as a discrete approximation version:}  
\begin{equation}
\label{lifapp} 
\textcolor{black}{u_i^{l,t} = \lambda u_{i}^{l-1, t-1} + \sum_{j} w_{ij} s_{j}^{l-1, t} - s_{i}^{l-1, t-1} \theta,} 
\end{equation}
\textcolor{black}{where $\theta$ represents the discharge threshold, and $u_i^{l,t}$ is the voltage of layer $l$ on step t.} 
\textcolor{black}{$s_i^{l}$ is used to indicate a neuron spike or not of layer $l$.} 
Specifically, $s_i^{t} = 1$, if the $u_i^t$ is larger than $\theta$, otherwise, $s_i^{t} = 0$. $u_i$ is the membrane potential of a neuron $i$.  $\lambda$ is the leaky constant for the membrane potential, which is less than 1. $w_{ij}$ denotes the weight connecting the neuron $i$ and its pre-synaptic neuron $j$.

To help the readers better understand the spiking neurons, we give a comparison between the widely used ANN neuron, IF, and LIF spiking neurons. 
\textcolor{black}{
The ANN neuron (as shown in Fig.~\ref{spikingNeuron} (a)) is widely used in the current deep learning network architectures, such as the ReLU (Rectified Linear Unit) activation layer which outputs the input directly if it's positive, and zero otherwise. It helps the model to learn complex patterns by introducing non-linearity. 
For the IF (Integrate-and-Fire) neuron in Fig.~\ref{spikingNeuron} (c), it integrates incoming signals until a certain threshold is reached, at which point it \textit{fires} or produces an output spike. The IF neuron captures the basic idea of neural activity despite its simplicity. 
For the LIF (Leaky Integrate-and-Fire) neuron, as shown in Fig.~\ref{spikingNeuron} (b), it introduces a leakage term to the integration process and this makes the accumulated charge decrease over time if it doesn't reach the firing threshold. Note that this leakage design mimics the biological reality that neurons gradually lose charge if not stimulated. 
In a nutshell, ReLU is great for traditional ANNs dealing with continuous data, while IF and LIF neurons are more specialized for spiking neural networks, capturing the temporal dynamics seen in biological neural systems. 
}

As a kind of biomimetic neural network, our proposed \textcolor{black}{SCNN} encoder can obviously reduce energy consumption. In each simulation step of the \textcolor{black}{SCNN} encoder, we extract and accumulate the voltage information after the 4$^{th}$, 6$^{th}$, 8$^{th}$ convolutional layer and get the $A1, A2, A3$ feature representations, as shown in Fig.~\ref{framework}. To achieve a more robust feature representation learning, we introduce two deconvolutional layers (i.e., T1 and T2 in Fig.~\ref{framework}) to augment and fuse the multi-scale voltage information. It is worth noting that such a hybrid SNN-ANN network architecture can achieve a better trade-off between energy consumption and overall performance.

\begin{figure*}
\center
\includegraphics[width=7in]{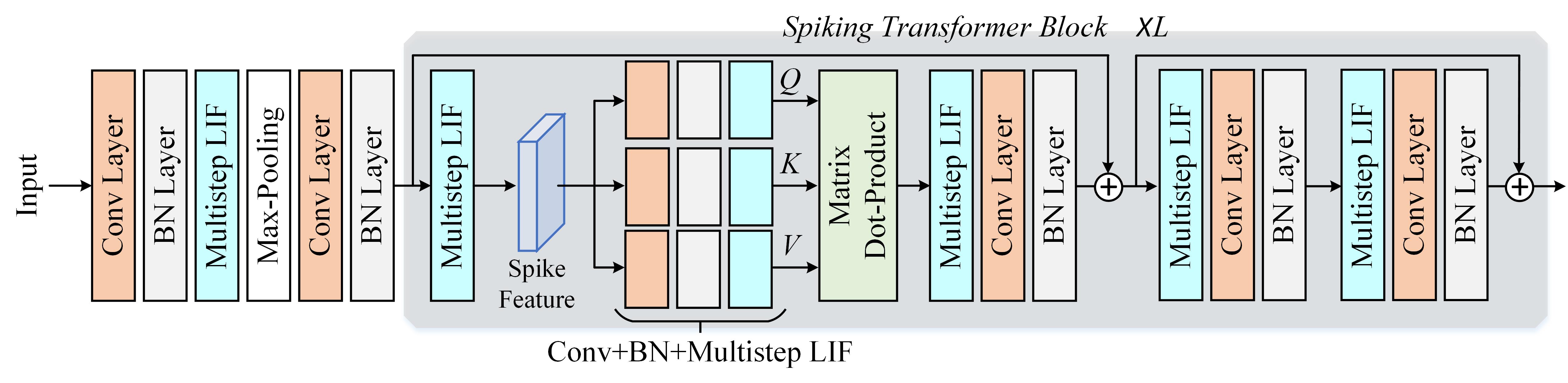}
\caption{An illustration of SpikingFormer block~\cite{zhou2023spikingformer}.} 
\label{SpikFormer}
\end{figure*}

\noindent  
\textbf{Memory Support Transformer. }
To handle the RGB modality, in this paper, we propose a novel memory support Transformer which \emph{recurrently} learns the spatial-temporal features via cross-attention between query and support frames. More in detail, we first adopt a pre-trained ResNet to obtain the feature embeddings $\mathcal{\hat{F}} = \{\hat{f_1}, \hat{f_2}, ..., \hat{f_N}\}$ of given frames $\mathcal{F} = \{f_1, f_2, ..., f_N\}$. Then, we split these feature embeddings into multiple clips. In our implementation, each clip contains the features corresponding to four frames. 
\textcolor{black}{
The first three feature embeddings and the output of the previous clip are treated as the support features and the last one is the query feature. The key idea is borrowed from few-shot learning, and the query frame in the few-shot learning denotes the samples needed to be \textcolor{black}{inferred}, and the support features denote the samples with labels which helps the inference of the query frame. This mechanism in our case will boost the feature learning of query frames based on temporal information provided by support features. 
}
\textcolor{black}{
To achieve this target, we introduce a GRU (Gated Recurrent Unit) network to learn the temporal information of support features $x$ and the detailed compute procedure can be summarized as: 
\begin{flalign} 
&~~r_t = \delta(W_{ir}x_t + b_{ir} + W_{hr}h_{(t-1)} + b_{hr}), \\
&~~z_t = \delta(W_{iz}x_t + b_{iz} + W_{hz}h_{(t-1)} + b_{hz}), \\
&~~n_t = tanh(W_{in}x_t + b_{in} + r_{t} \odot (W_{hn}h_{(t-1)} + b_{hn})), \\ 
&~~h_t = (1-z_t) \odot n_t + z_t \odot h_{(t-1)},   
\end{flalign}
where $W_{ir}, W_{hr}, W_{iz}, W_{hz}, W_{in}, W_{hn}$ are learnable weight parameters, $b_{ir}, b_{hr}, b_{iz}, b_{hz}, b_{in}, b_{hn}$ are bias in the neural networks. $h_{(t-1)}$ and $h_{t}$ is the hidden feature representation at the time $t-1$ and $t$, respectively. $tanh$ is the activation layer, $\delta$ is the sigmoid function, $\odot$ is the Hadamard product, $r_t, z_t, n_t$ are the reset, update, and new gates, respectively. 
} 
As our MST addresses the temporal features in a recurrent manner, 
\textcolor{black}{the output of cross-attention at the previous step is also considered in the current step via concatenate operation.} 
\textcolor{black}{
Given the output of GRU network, we adopt the cross-attention mechanism to fuse the support features and the query features. 
}
In our case, the query $Q$ is the last embedding in a clip, \textcolor{black}{key $K$ and value $V$ \textcolor{black}{are} the output of GRU network.} 
Mathematically speaking, 
\begin{equation}
\label{} 
CrossAttn = Softmax(\frac{QK^{T}}{\sqrt{c}}) V 
\end{equation} 
where $c$ is the feature dimension. The outputs of the memory support Transformer will be concatenated with SCNN features for the final prediction.


\noindent 
\textbf{Multi-modal Bottleneck Fusion. }
In this paper, we aggregate the RGB frames and event streams for more accurate pattern recognition by  employing a  bottleneck fusion mechanism~\cite{nagrani2021bottlenecks}. Specifically, we randomly initialize a feature map $Z \in \mathbb{R}^{16 \times 60 \times 60}$ and concatenate with the output ($ \in \mathbb{R}^{16 \times 60 \times 60}$) of SCNN branch. Then, we propose a deformable convolutional block to fuse these features, as shown in Fig.~\ref{framework}, which mainly contains a 2D convolutional layer and 4 deformable convolutional layers. The max-pooling operation is also used after the first and second convolutional layers. Group normalization~\cite{wu2018groupnormalization} and ReLU activation layers are also used in this module. The output feature of this module is $32 \times 14 \times 14$. We split this feature into two parts, i.e., the event representation $16 \times 14 \times 14$ and the bottleneck features $16 \times 14 \times 14$. \textcolor{black}{The bottleneck features} will be fed into the memory support Transformer network for modality interactive learning.

The output feature of the memory support Transformer and the event representation will be concatenated and flattened into a feature representation (7232-D). Then, a prediction head with two fully connected layers is proposed for recognition. The dimensions of the first and second FC layers are set to 4096 and the number of classification categories respectively.

\noindent 
\textbf{Loss Function. } \label{optimization}
In this paper, we adopt the cross-entropy loss function to measure the distance between our prediction $\bar{y}$ and the ground truth $y$. To be specific, the ground truth $y$ is represented as a $one\text{-}hot$ vector. The loss function can be written as: 
\begin{equation}
\label{CELoss} 
Loss = - [y log \bar{y} + (1-y) log (1-\bar{y})]. 
\end{equation}

\subsection{Dual-Transformer based Version}~\label{networkArch} 
To achieve a higher recognition performance, in this paper, we also exploit Spiking Transformer Network~\cite{zhou2023spikingformer} to process the event stream. As shown in Fig.~\ref{SpikFormer}, the input is first encoded using the convolutional module, which contains Conv+BN layer, multistep LIF neuron, and max-pooling layer. Note that the multistep LIF neuron is adopted as the activation function. The output will be binary spike features after the multistep LIF neuron. Then, we adopt Conv+BN+Multistep LIF module to project it into query (\emph{Q}), key (\emph{K}), and value (\emph{V}) tokens and conduct matrix dot-product operation similar to Transformer networks. Then, a couple of LIF+Conv+BN modules and residual connections are adopted for feature enhancement.

In our implementation, the output token and its dimension are $336 \times 256$. We initialize 64 learnable bottleneck tokens and concatenate them with the output tokens from the spiking Transformer, i.e., $400 \times 256$. Two Transformer blocks are adopted to process the tokens and the output will be split into two parts. The first part is used for interaction with the memory support Transformer, and the second part will be fused with the enhanced RGB features for final recognition. Our experiments show that such dual-Transformer network architecture can obtain better recognition performance.

\section{Experiments} \label{Experiments} 

\subsection{Dataset and Evaluation Metric} \label{datasetPoker}
In this paper, we conduct extensive experiments on two event-based classification datasets, including \textbf{HARDVS}~\cite{wang2022hardvs}, and our newly proposed \textbf{PokerEvent} dataset. HARDVS dataset is collected using a DVS346 event camera with a focus on human activity recognition, such as \emph{running, holding an umbrella, taking on shoes}, etc. The dataset contains 107646 RGB-Event samples and they split these videos into training, validation, and testing subset with 64526, 10734, 32386, samples, respectively.

\begin{figure} 
\center
\includegraphics[width=3.3in]{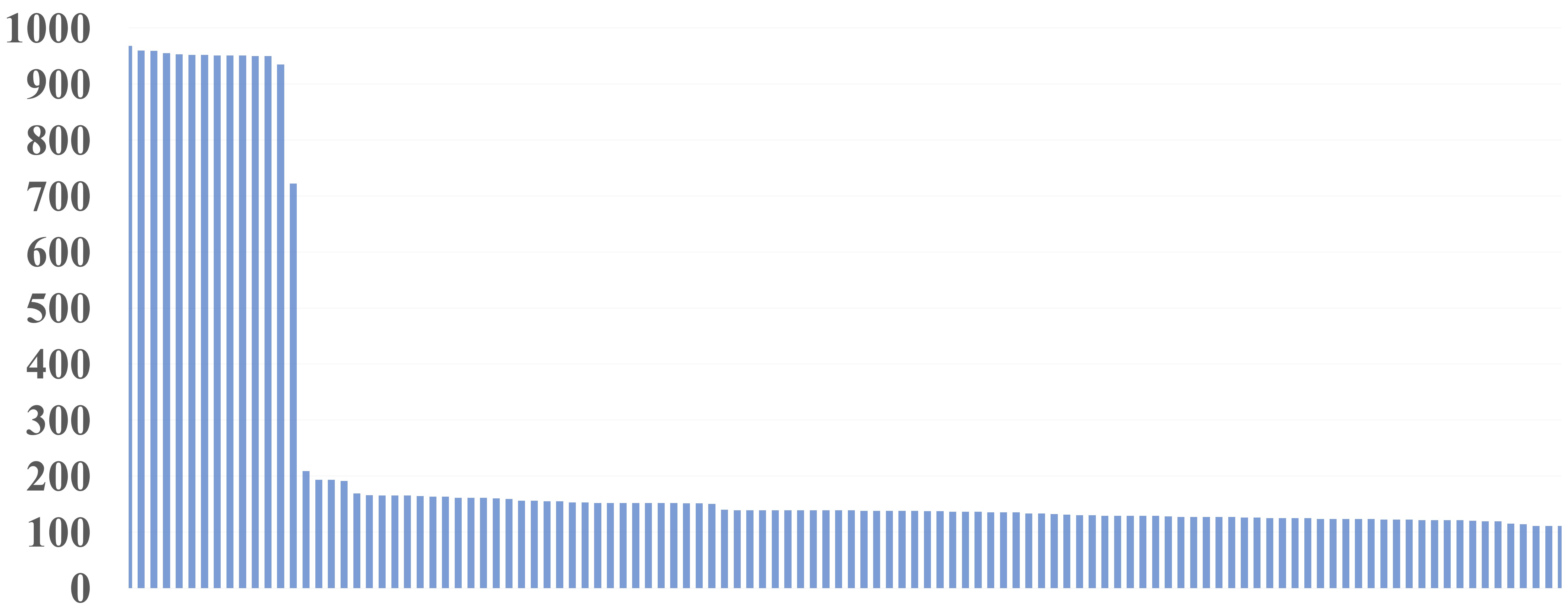}
\caption{Distribution of the number of RGB-Event samples of each category in our newly proposed PokerEvent dataset.}
\label{DistributionPokerEvent} 
\end{figure}

\begin{figure*}
\center
\includegraphics[width=7in]{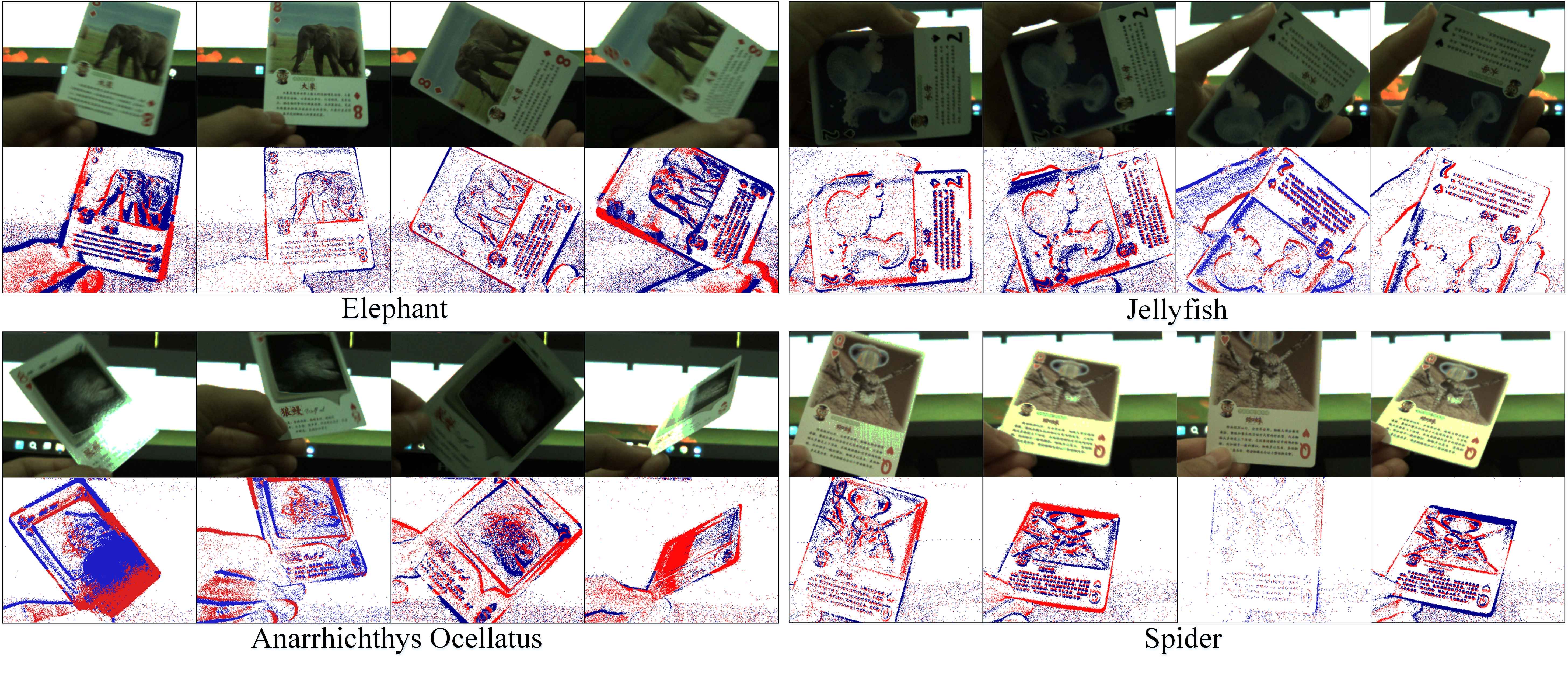}
\caption{Some representative examples of our newly proposed PokerEvent dataset.} 
\label{Sample_img}
\end{figure*}

A new RGB-Event pattern recognition benchmark dataset is proposed in this paper, termed PokerEvent. As the name shows, the target object of this dataset is the character patterns in the poker card. It contains 114 classes, 16216/2687/8199 training/validation/testing samples (the spatial resolution is $346 \times 260$) recorded using a DVS346 event camera. The distribution information of the number of RGB-Event samples of each category can be found in Fig.~\ref{DistributionPokerEvent}. We also give some visualization of representative samples in Fig.~\ref{Sample_img}.

For the evaluation of experimental results, we select the \textbf{top-1} and \textbf{top-5} accuracy as the evaluation metrics.

\subsection{Implementation Details}
In our experiments, the batch size is set as 4, and Adam optimizer~\cite{kingma2014adam} is selected for the training with an initial learning rate of 0.001. The simulation steps of \textcolor{black}{SCNN} are fixed as 16, which is equal to the number of frames in one video sample. For the loss function, we select the widely used binary cross-entropy (BCE) loss to measure the distance between the ground truth and our predictions. The classification label is embedded into the corresponding one-hot feature vector. Our model contains 336.48M parameters and can run at 0.22 seconds for an RGB-Event sample. Our code is implemented with Python based on PyTorch~\cite{paszke2019pytorch}. All the experiments are conducted on a server with RTX3090 GPUs.

\subsection{Comparison with Other SOTA Models} 

In this section, we report the classification results of ours and other state-of-the-art models on the HARDVS and PokerEvent datasets. 

\noindent
$\bullet$ \textbf{Results on PokerEvent dataset.~} 
In this work, we train and test multiple recognition models for future works to compare, including the CNN-based model C3D~\cite{tran2015c3d}, TSM~\cite{lin2019tsm}, ACTION-Net~\cite{wang2021actionnet}, TAM~\cite{liu2021tam}, X3D~\cite{feichtenhofer2020x3d}, and Transformer-based model V-SwinTrans~\cite{liu2021videoSwin}, TimeSformer~\cite{liu2021videoSwin}, MVIT~\cite{li2022mvitv2}, etc. As these models are specifically developed for RGB-based recognition, our proposed task needs to fuse the RGB frames and event streams. Thus, we train these compared methods using concatenated RGB frames and event images.

As shown in Table~\ref{pokerResults}, our proposed model achieves $53.19\%$ on the top-1 accuracy on the  PokerEvent dataset, which is better than C3D~\cite{tran2015c3d} and X3D model~\cite{feichtenhofer2020x3d}. Note that the C3D adopts the 3D convolutional neural networks to learn the spatial-temporal features which have been widely validated in its effectiveness. The X3D is developed by progressively expanding the 2D network across various axes, including temporal duration, frame rate, etc. Our result is comparable with ACTION-Net, TAM, and Video-SwinTransformer, which are 54.29, 53.65, and 54.17 on top-1 accuracy, respectively. This demonstrates that our proposed model achieves comparable performance with existing state-of-the-art recognition algorithms.

\begin{table} 
\scriptsize         
\center 
\caption{Results (Top-1) on the newly proposed PokerEvent dataset.}    
\label{pokerResults}
\scalebox{1}{
\begin{tabular}{l|l|c|cccc} 
\hline \toprule [0.5 pt]
\textbf{Algorithm} &\textbf{Publish} &\textbf{Backbone} &\textbf{Event+Frame}    \\
\hline 
C3D~\cite{tran2015c3d} &ICCV-2015       &3D-CNN   &51.76              \\
\hline 	
TSM ~\cite{lin2019tsm} &ICCV-2019         &ResNet-50       &55.43     \\
\hline 	
ACTION-Net~\cite{wang2021actionnet} &CVPR-2021     &ResNet-50   &54.29 \\
\hline 	
TAM \cite{liu2021tam} 	&ICCV-2021         &ResNet-50       &53.65      \\
\hline 
V-SwinTrans \cite{liu2021videoSwin}   &CVPR-2022         &Swin Transformer      &54.17          \\
\hline 
TimeSformer \cite{bertasius2021TimeSformer}  &ICML-2021         &ViT      &55.69        \\
\hline 
X3D  \cite{feichtenhofer2020x3d}  &CVPR-2020         &ResNet       & 51.75           \\
\hline 
MVIT  \cite{li2022mvitv2}  &CVPR-2022         &ViT       &55.02       \\
\hline 
SAFE \cite{li2023semanticaware}  &arXiv2023         &ViT       & 57.64           \\
\hline 
TSCFormer \cite{wang2023unleashing}  &arXiv2023         &TSCFormer       &\textcolor{black}{\textbf{57.70}}       \\
\hline  
Ours (SCNN-MST)  &-         &SCNN-Former       &53.19       \\
\hline  
Ours (Spikingformer-MST)  &-         &SNN-Former       &54.74       \\
\hline \toprule [0.5 pt]
\end{tabular} }   
\end{table}


\noindent
$\bullet$ \textbf{Results on HARDVS dataset~\cite{wang2022hardvs}.~} 
The HARDVS benchmark dataset provides both RGB frames and event streams for human activity recognition. The authors only report the recognition results based on event streams. In this paper, we provide the experimental results on RGB frames, and multimodal settings (i.e., RGB frames and event streams). 
For the event stream based recognition, we train and test our proposed SCNN sub-network for comparison. As shown in Table~\ref{HARDVSResults_SM}, we can find that our model achieves 49.02 on the top-1 recognition results. This result is better than the fully ANN-based model ACTION-Net (46.85) and X3D (45.82), which validates the good performance of our SCNN sub-network. The performance of SCNN is also comparable to C3D, TAM, and TimeSformer. These results demonstrate that our SCNN network achieves comparable or even better results than the state-of-the-art models.
For the RGB frames, we train and test our proposed memory support Transformer (MST) network and compare it with other SOTA models. From Table~\ref{HARDVSResults_SM}, it is easy to find that our MST achieves 48.17 on the top-1 accuracy, this result is comparable to or even better than ESTF, X3D, ACTION-Net, and C3D. These experimental results validated that our MST can achieve good results for RGB-based recognition.

When combining the RGB frames and event streams, we can find that better results can be achieved by comparing with a single modality-based version. For example, by comparing the results reported in Table~\ref{HARDVSResults_SM} and Table~\ref{HARDVSResults_MM}, the top-1 accuracy of the C3D model can be improved to 50.88, while it achieves 50.52 and 49.94 on the Event and RGB modality only. TimeSformer improves the single modality from 50.77/50.05 (Event/RGB) to 51.57 on the top-1 accuracy. However, not all the recognition models are improved with a simple concatenate fuse method. For instance, the ESTF model is originally developed for event-based recognition only. Its result drops from 51.22/48.03 (Event/RGB) to 49.93 on the top-1 accuracy. Our experimental results are improved with the fusion of RGB frames and event streams with the help of multi-modal bottleneck fusion module. Specifically, our results are improved to 49.40/55.55 on the top-1 and top-5 accuracy. These results fully validated the effectiveness of our proposed RGB-Event recognition framework.

\begin{figure*}
\centering
\includegraphics[width=7in]{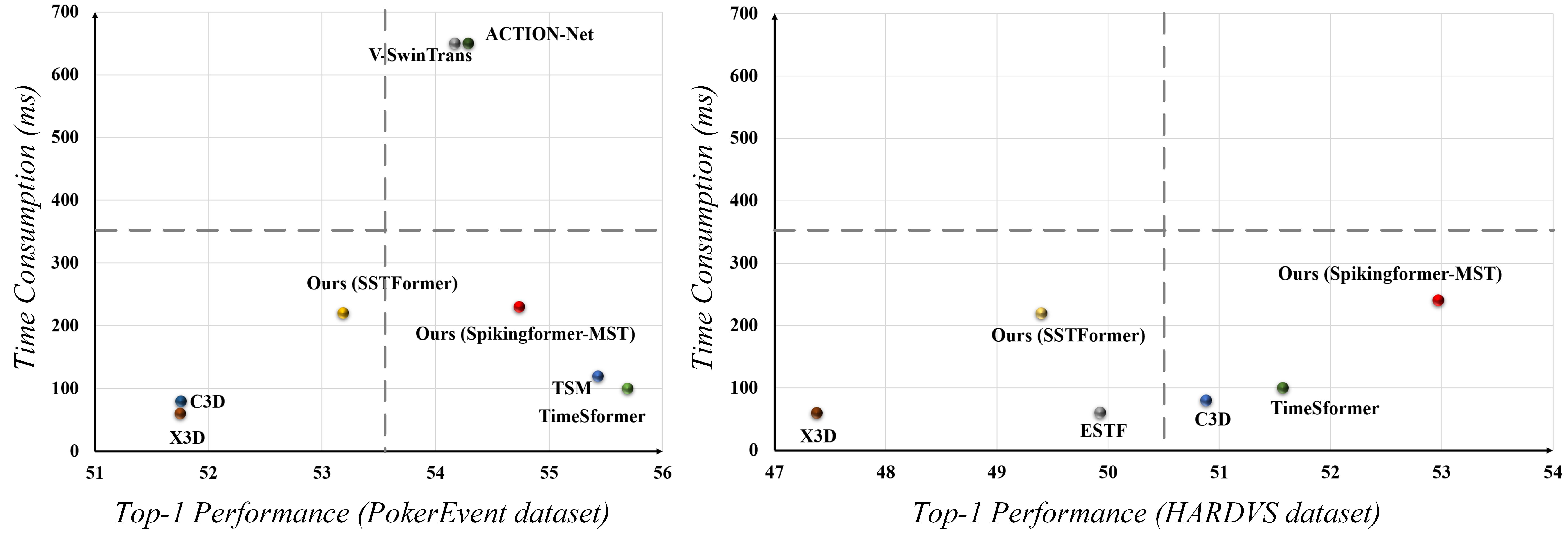}
\caption{
\textcolor{black}{An illustration of trade-off between the recognition performance and time consumption on PokerEvent and HARDVS.} 
} 
\label{fig:timeacctradeoff}
\end{figure*}

\begin{table} 
\small          
\center 
\caption{ \textcolor{black}{Results (Top-1 | Top-5) on the HARDVS dataset (Single Modality).} }     
\label{HARDVSResults_SM}
\scalebox{1}{
\begin{tabular}{l|c|c|ccc} 
\hline \toprule [0.5 pt]
\textbf{Algorithm} &\textbf{Backbone} &\textbf{Event} &\textbf{RGB}   \\
\hline 
C3D~\cite{tran2015c3d} &3D-CNN   &50.52 $|$ 56.14    &49.94 $|$ 55.05         \\
\hline 	
TSM ~\cite{lin2019tsm} &ResNet-50       &\textcolor{black}{\textbf{52.63}}   $|$ 60.56 &\textcolor{black}{\textbf{51.70}} $|$ 58.87     \\
\hline 	
ACTION-Net~\cite{wang2021actionnet} &ResNet-50   &46.85 $|$ 56.19   &48.05 $|$ 54.24 \\
\hline 	
TAM \cite{liu2021tam} &ResNet-50       &50.41 $|$ 57.99 &50.31 $|$ 56.91     \\
\hline  
TimeSformer \cite{bertasius2021TimeSformer}   &ViT      &50.77 $|$ 58.70 &50.05 $|$ 55.54          \\
\hline 
X3D  \cite{feichtenhofer2020x3d}    &ResNet       & 45.82 $|$ 52.33 &48.31 $|$ 51.37         \\
\hline 
ESTF  \cite{wang2022hardvs}    &ResNet-18       & 51.22 $|$ 57.53 &48.03 $|$ 51.01         \\
\hline \toprule [0.5 pt] 
SCNN    &SCNN    &49.02 $|$ 53.62   &-   \\  
\hline 
Spikingformer    &SNN    &51.87 $|$ 58.67    &-     \\ 
\hline 
MST     &Former    &-    &48.17 $|$ 51.39    \\ 
\hline \toprule [0.5 pt]
\end{tabular} }   
\end{table}


\begin{table} 
\small        
\center 
\caption{ \textcolor{black}{Results (Top-1 | Top-5) on the HARDVS dataset (Event+RGB).} }     
\label{HARDVSResults_MM}
\scalebox{1}{
\begin{tabular}{l|c|cc} 
\hline \toprule [0.5 pt]
\textbf{Algorithm} &\textbf{Backbone} &\textbf{Event+RGB}   \\
\hline 
C3D~\cite{tran2015c3d} &3D-CNN     &50.88 $|$ 56.51       \\
\hline 	
TimeSformer \cite{bertasius2021TimeSformer}   &ViT      &51.57 $|$ 58.48          \\
\hline 
X3D  \cite{feichtenhofer2020x3d}    &ResNet         &47.38 $|$ 51.42        \\
\hline 
ESTF  \cite{wang2022hardvs}    &ResNet-18        &49.93 $|$ 55.77        \\
\hline 
Ours (SCNN-MST)  &SCNN-Former        &49.40 $|$ 55.55    \\
\hline 
Ours (Spikingformer-MST) &SNN-Former        &\textcolor{black}{\textbf{52.97}} $|$ 60.17        &            \\  
\hline \toprule [0.5 pt]
\end{tabular} }   
\end{table}

\subsection{Ablation Study} 
In this subsection, we will first analyze the key components in our model to check their contributions to the final results. Then, we will give an analysis of the number of video frames, different spiking neurons, and energy consumption, etc. 

\noindent
\textbf{Component Analysis.~}
Our proposed model contains the memory support Transformer (MST), Spiking Convolutional Neural Network (\textcolor{black}{SCNN}), Multi-modal Bottleneck Fusion (MBF) modules, etc. We will report their recognition performance separately on the HARDVS and PokerEvent datasets. As shown in Table~\ref{CAResults}, we can find that the MST module achieves 52.14 and 48.17 on the top-1 accuracy on the PokerEvent and HARDVS datasets. The SCNN module obtains $37.61, 49.02$ on the two metrics. Note that the two modules are designed for single-modality processing, in other words, we feed the RGB frames/event streams into the MST and SCNN modules only. When both modalities are used, the overall results can be improved to $53.23$ and $49.13$, which fully demonstrate the effectiveness of combining the RGB frames and event streams for pattern recognition. Our results can be further enhanced with the multimodal bottleneck fusion (MBF) module, i.e., $53.80, 49.40$ on the top-1 and top-5 metrics. This result demonstrates that the MBF module also contributes to our final recognition performance.

\begin{table} 
\center
\small     
\caption{Component Analysis on the PokerEvent and HARDVS Dataset.} 
\label{CAResults} 
\begin{tabular}{c|ccc|cc} 		
\hline \toprule [0.5 pt] 
\textbf{No.}  & \textbf{MST} &\textbf{SCNN} &\textbf{MBF}  &\textbf{PokerEvent}  &\textbf{HARDVS} \\
\hline 
1 &\CheckmarkBold           &         &         &52.14   &48.17              \\
2 &   &\CheckmarkBold       &         &37.61    &49.02             \\
3 &\CheckmarkBold   &\CheckmarkBold    &        &53.23     &49.13             \\
4 &\CheckmarkBold   &\CheckmarkBold    &\CheckmarkBold   &53.80     &49.40              \\
\hline \toprule [0.5 pt]
\end{tabular}
\end{table}

\begin{figure}
\center
\includegraphics[width=3.5in]{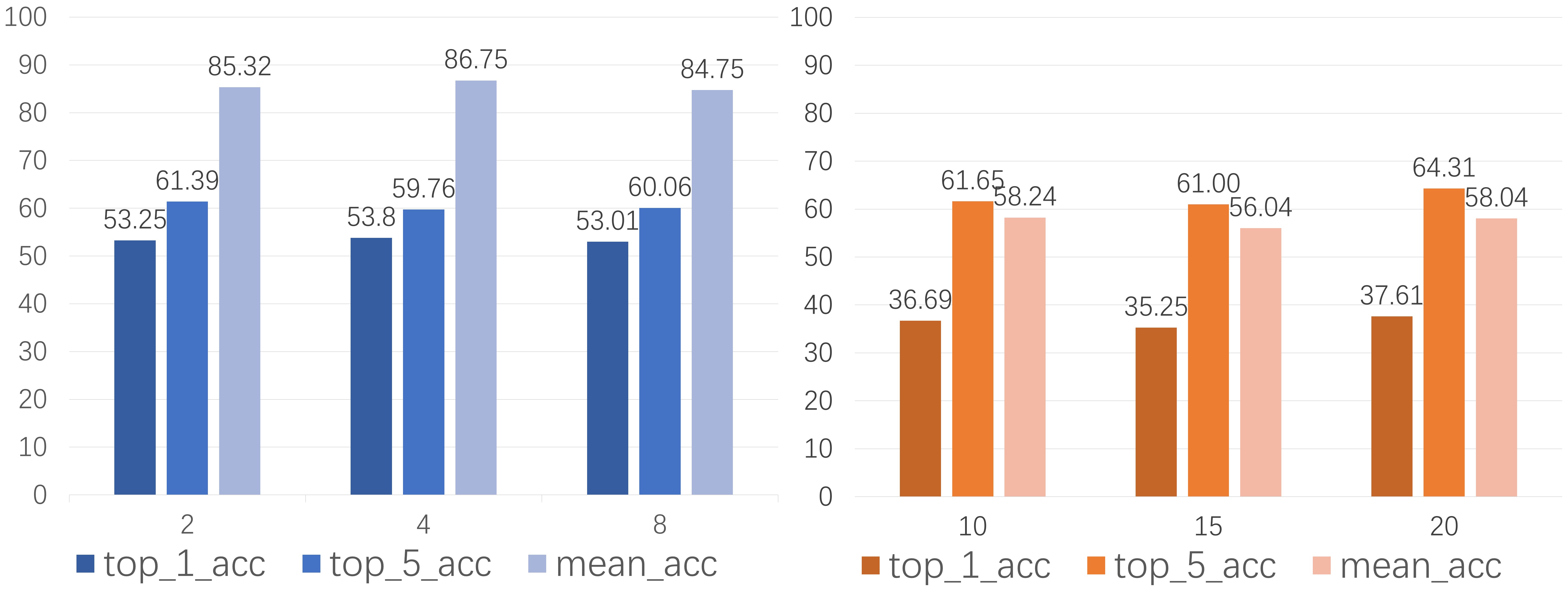}
\caption{(left) Different number of divided clips for Memory Support Network; (right) Different number of divided fragments of event streams on the PokerEvent dataset.}  
\label{clip_numberIMG}
\end{figure}

\begin{figure}
\center
\includegraphics[width=3.5in]{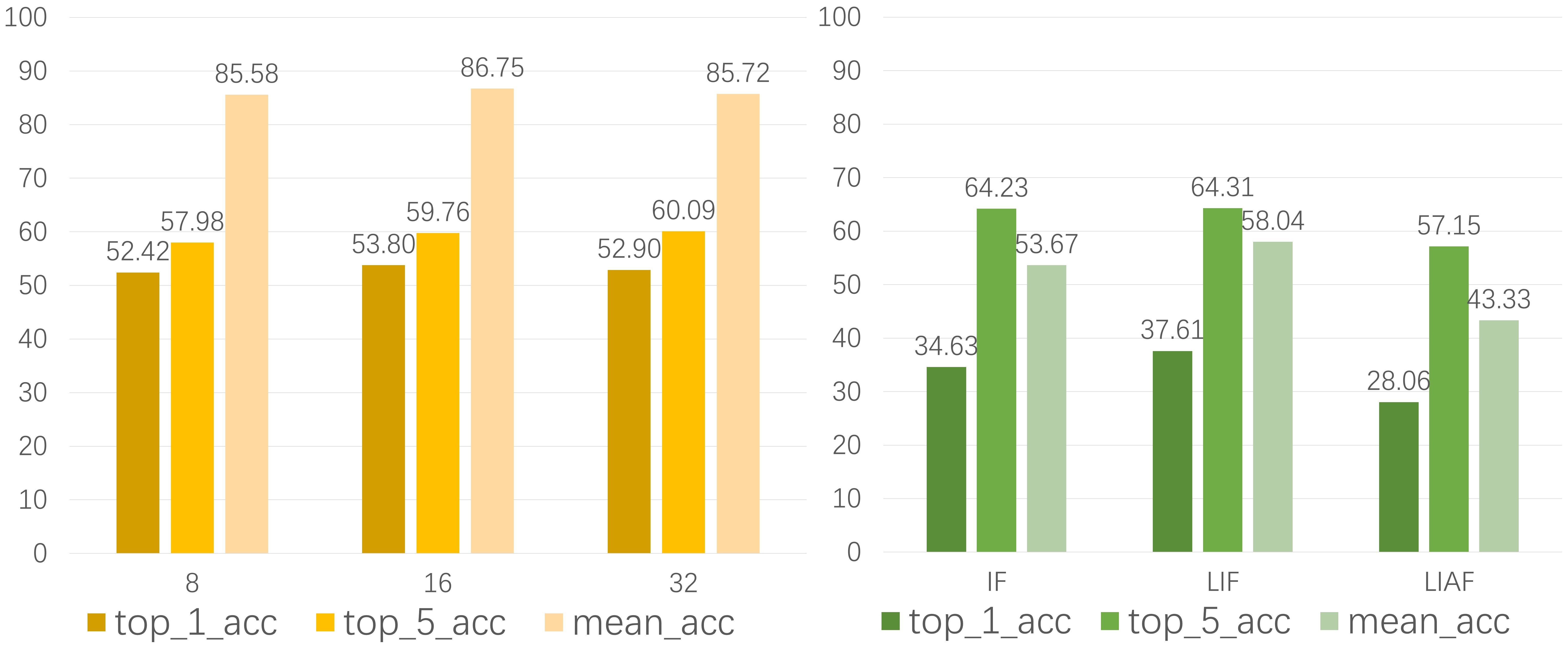}
\caption{ (left) Different dimensions of bottleneck feature maps; (right) Different spiking neurons, i.e., IF, LIF, and LIAF. }  
\label{dim_snn}
\end{figure}

\noindent
\textbf{Analysis on Number of Video Frames.~}  
In our implementation, the raw RGB frames are embedded into feature representations and divided into multiple clips to learn the spatial-temporal features. In this part, we will test the influence of the input frames by setting them to 2, 4, and 8 clips. Each corresponding clip contains 8, 4, and 2 frames, respectively. We can find that when the clips of the input frame are set to 4, better results can be achieved. Specifically, the top-1 accuracy, top-5 accuracy, and mean accuracy are 53.8, 59.76, and 86.75, respectively. Top-1/Top-5 accuracy is the proportion of all samples that are correctly predicted. In contrast, the mean accuracy first calculates the accuracy by category and then takes the average of these accuracies.


\noindent
\textbf{Analysis on Different Spiking Neurons.~} 
The spiking neurons play an important role in \textcolor{black}{SCNN}, in this paper, we test the following three neurons, i.e., IF~\cite{burkitt2006IFneuron}, LIF~\cite{wu2018lif}, and LIAF~\cite{wu2021liaf}. As illustrated in the right part of Fig.~\ref{dim_snn}, we can find that the LIF neuron achieves better results than the other two neurons, i.e., 37.61 and 64.31 on the event streams of the PokerEvent dataset.

\begin{figure} 
\center
\includegraphics[width=3.5in]{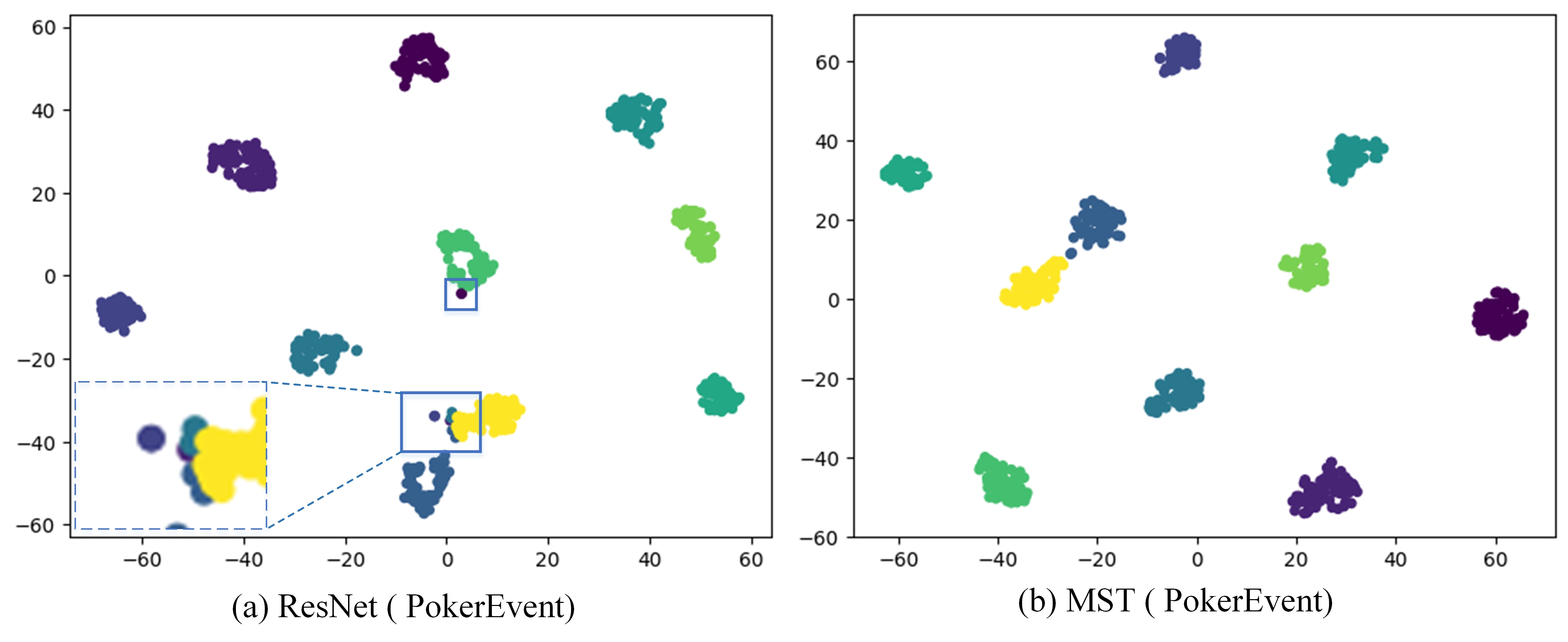}
\caption{Visualization of feature distribution of ResNet18 and MST on PokerEvent.} 
\label{RES_MST}
\end{figure}

\begin{figure*} 
\center
\includegraphics[width=7in]{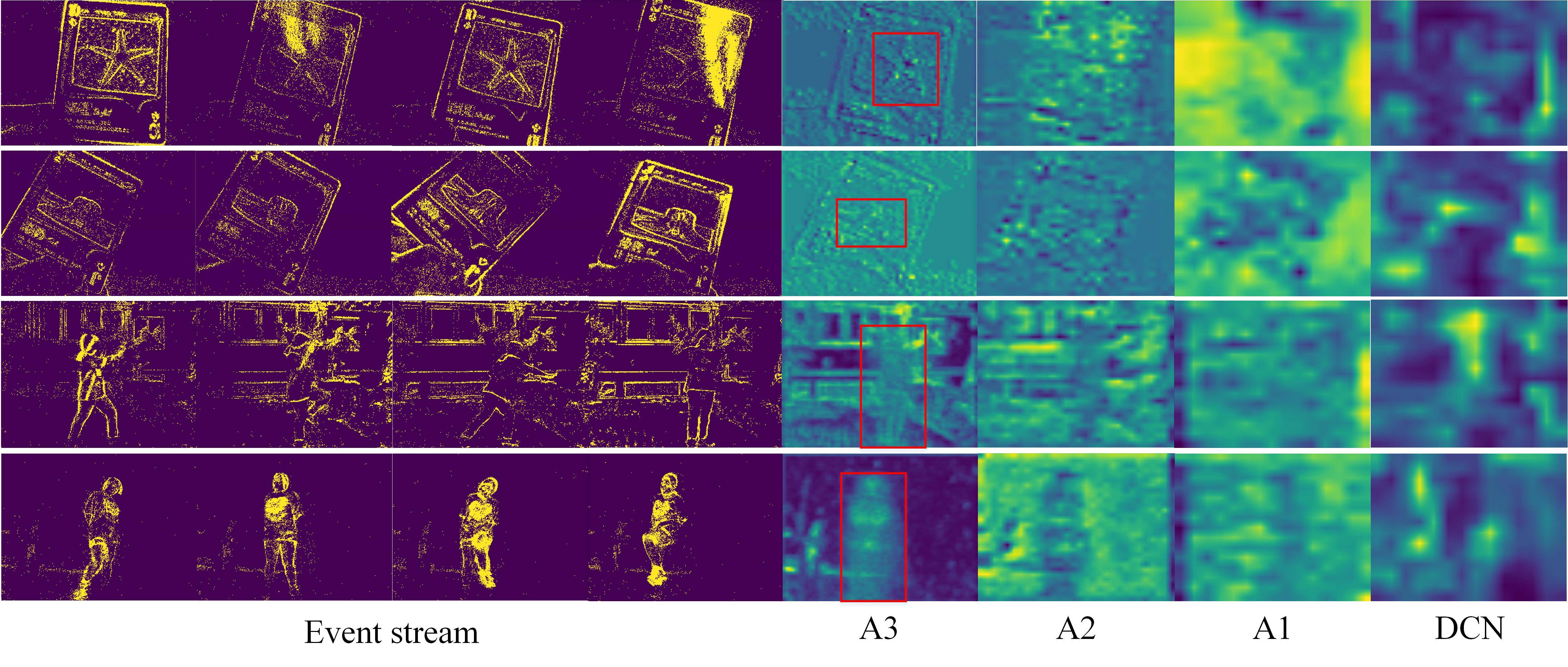}
\caption{Visualization of event stream and feature maps learned using SCNN and DCN module.} 
\label{SCNN_feature_map}
\end{figure*}


\noindent
\textbf{Analysis on Energy Consumption. } 
As is known to all, the SNNs cost much lower energy compared with corresponding similar ANNs~\cite{zhu2022eventsnn, rathi2021dietsnn}. Because each operation in ANN computes a dot product involving one floating-point (FP) multiplication and one FP addition (MAC), meanwhile, each operation in SNN is only one FP addition due to binary spikes, as noted in~\cite{rathi2021dietsnn}. 
As we know, the number of ANN operations (MAC) in a convolutional layer can be defined as: 
\begin{equation}
\label{OP_ANN} 
\#OP_{ANN} = k_w \times k_h \times c_{in} \times h_{out} \times w_{out} \times c_{out}
\end{equation}
where $k_w, k_h$ are the width and height of the kernel, $c_{in}, c_{out}$ are the number of input and output channels of feature maps, $h_{out}, w_{out}$ are the height and width of the output feature map. 
Correspondingly, the number of SNN operations is: 
\begin{equation}
    \label{OP_SNN} 
    \#OP_{SNN} = SpikeRate_l \times \#OP_{ANN} 
\end{equation}
where $SpikeRate_l$ is all the spikes in layer $l$ over all timesteps averaged over the number of neurons of layer $l$. We can deem the number of operations for SNN and ANN is the same, if the spike rate is 1, i.e., every neuron fired once.

To give a more intuitive explanation, in this section, we will compute the Energy ConsumPtion (ECP) via the following formula which can reflect the improvement over the corresponding ANN version (i.e., replacing the spiking neurons with the widely used activation function ReLU):  
$ECP_{SNN} = ECP_{ANN} * SpikeRate_l, $
where the $ECP_{SNN}$ and $ECP_{ANN}$ denotes the energy cost of the $l$-th layer of SNN and ANN network, respectively. Note that the energy cost/operation for ANNs and SNNs is computed in 45 nm CMOS technology by following~\cite{rathi2021dietsnn}. The statistics show that each operation in ANN and SNN cost about $4.6 pJ$ and $0.9 pJ$, respectively~\cite{rathi2021dietsnn, horowitz20141}.

In our case, the resolution of input is $240 \times 240$, and the channel number of eight SNN layers and two deconvolutional layers are [64, 64, 128, 128, 256, 256, 512, 512] and [256, 128], with the adjacent two SNN layers having the same channel dimension. We utilize a convolutional kernel with size $3 \times 3$ for all eight layers and $4 \times 4$ for the deconvolutional layers. The size of SNN output feature maps is [$240 \times 240, 240 \times 240, 120 \times 120, 120 \times 120, 60 \times 60, 60 \times 60, 30 \times 30, 30 \times 30$] and two deconvolutional feature maps are [$30 \times 30$, $60 \times 60$]. Therefore, the number of neurons of spiking convolutional layers and two deconvolutional layers is 12,076,646,400 and 3774873600, respectively, based on Eq. (\ref{OP_ANN}). 
In all our 16 simulation steps, the spike emitted an average total of 21,971,781 times for 100 randomly selected samples. Therefore, the spike rate is about $21,971,781/(12,076,646,400*16) \approx 0.01137\%$. The relative improvement of SNN (0.9*[(12,076,646,400*16*0.0001137) + 3774873600]) over ANN (4.6 * (12,076,646,400*16 + 3774873600)) is about $\times 265$. 

\textcolor{black}{In the Table~\ref{param_time}, we represent the original model of spiking neurons with SCNN and the model of spiking neurons with ReLU activation function instead of Spiking Neurons with ANN. As shown in the table, Params represents the number of parameters for the model, and Time represents the time the model worked on the test set. The number of parameters in the ANN model is 638M, and the time required to run the entire test set is 58 minutes and 25 seconds. The number of parameters for the SCNN model is 638M, and the time to run the entire test set is 49 minutes and 46 seconds. Compared with ANN networks, SCNN is faster with the same number of parameters.}

\begin{table}
\center
\small   
\caption{\textcolor{black}{Comparison of the number of model parameters and test time.}} 
\label{param_time}
\begin{tabular}{l|c|c|c}
\hline 
\textbf{No.}   &\textbf{Method} &\textbf{Params}  &\textbf{Time}  \\
\hline
\textbf{01}  &\ SCNN     &\ 638M  &\ 49min46s   \\ 
\textbf{02}  &\ ANN     &\ 638M  &\ 58min25s   \\
\hline
\end{tabular}
\end{table}








\noindent
\textbf{Analysis on Different Event Fragments and CLIP.~} 
In this paper, we divide the event streams into multiple segments and extract the spatial-temporal features using spiking neural networks. We test different settings to check their influence on the final results. As illustrated in the right part of  Fig.~\ref{clip_numberIMG}, the top-1/top-5 accuracy is 36.69/61.65, 32.25/61.00, 37.61/64.31 when dividing the event streams into 10, 15, and 20 segments, respectively. In MST, we obtained different results by setting different numbers of clips. It can be seen that when the number of clips is 2, 4, and 8, the top-1 accuracy is 53.25, 53.8, and 53.01, respectively, 
\textcolor{black}{as shown in the left part of Fig.~\ref{clip_numberIMG}.}

\noindent
\textbf{Analysis on Different Dimensions of Bottleneck Feature.~}  
After the \textcolor{black}{SCNN} encoder, we integrate a bottleneck feature map into the decoding phase to achieve interactive learning between the RGB frame and event stream. In our implementation, the spatial resolution of this feature map is fixed, while its dimension is variable. We set it to 8, 16, and 32 and report their results in the left part of Fig.~\ref{dim_snn}. It is easy to find that the top-1/top-5 accuracy is 52.42/57.98, 53.80/59.76, and 52.90/60.09, respectively. The top-1 accuracy will be the best (53.80) when the dimension is 16, while the top-5 accuracy is the best (60.09) if 32-d bottleneck feature maps are used.

\noindent 
\textbf{Analysis on Dual-Transformer based Recognition Model.~}  
As shown in Table~\ref{pokerResults} and Table~\ref{HARDVSResults_MM}, the proposed Spikingformer-MST achieves 54.74 (top-1) on PokerEvent dataset and 52.97$|$60.17 (top-1$|$top-5) on HARDVS dataset, respectively. These results are comparable to or even better than recent SOTA models which fully validated the effectiveness of our proposed SNN-Former framework for RGB-Event based pattern recognition. Note that as shown in Table~\ref{HARDVSResults_SM}, when only event data is used for recognition, the Spikingformer achieves 51.87$|$58.67 (top-1$|$top-5) on the HARDVS dataset. It is inferior to the multimodal version which demonstrates the effectiveness of fusing RGB and event streams for recognition.

\begin{figure*}[!htp]
\center
\includegraphics[width=7in]{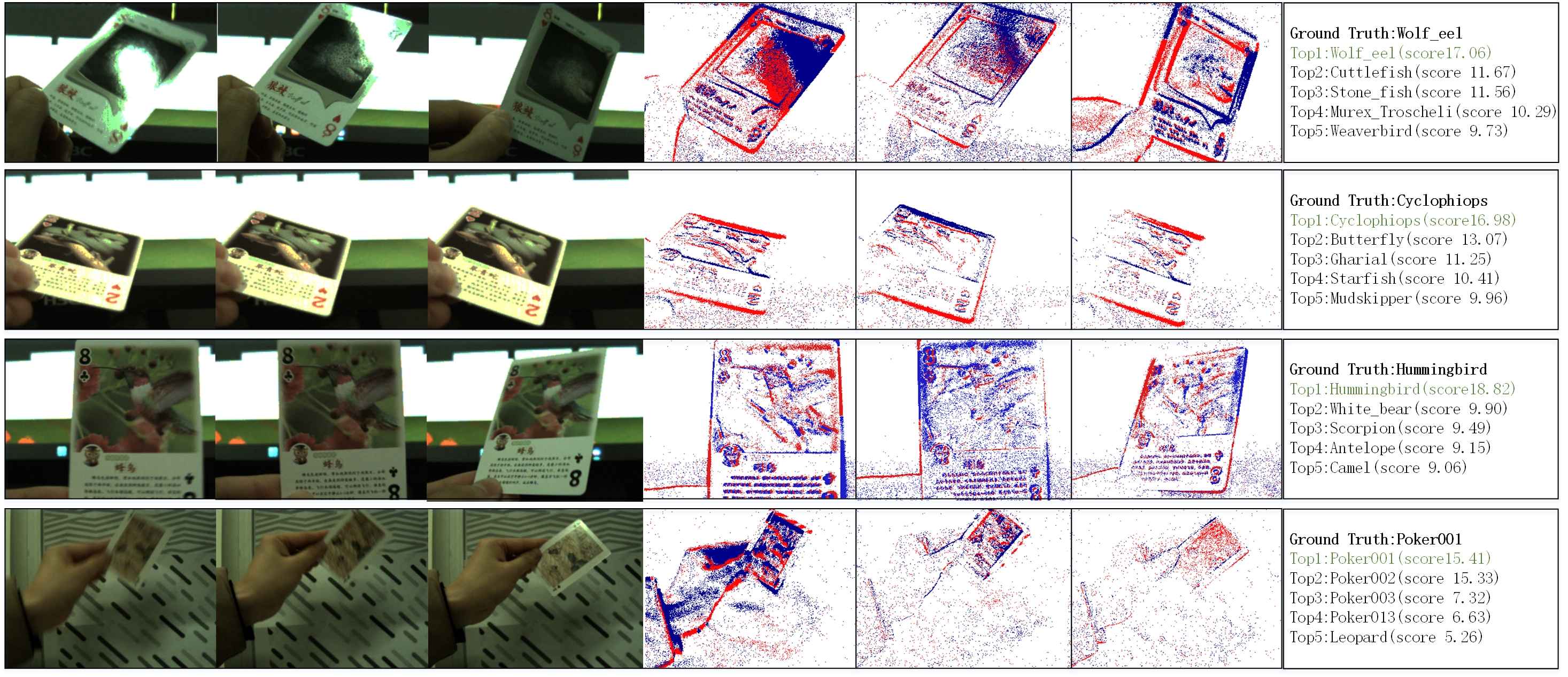}
\caption{Visualization of the top-5 predicted results on the PokerEvent dataset.} 
\label{pokerevent_top5.jpg}
\end{figure*}

\begin{figure*}[!htp]
\centering 
\includegraphics[width=7in]{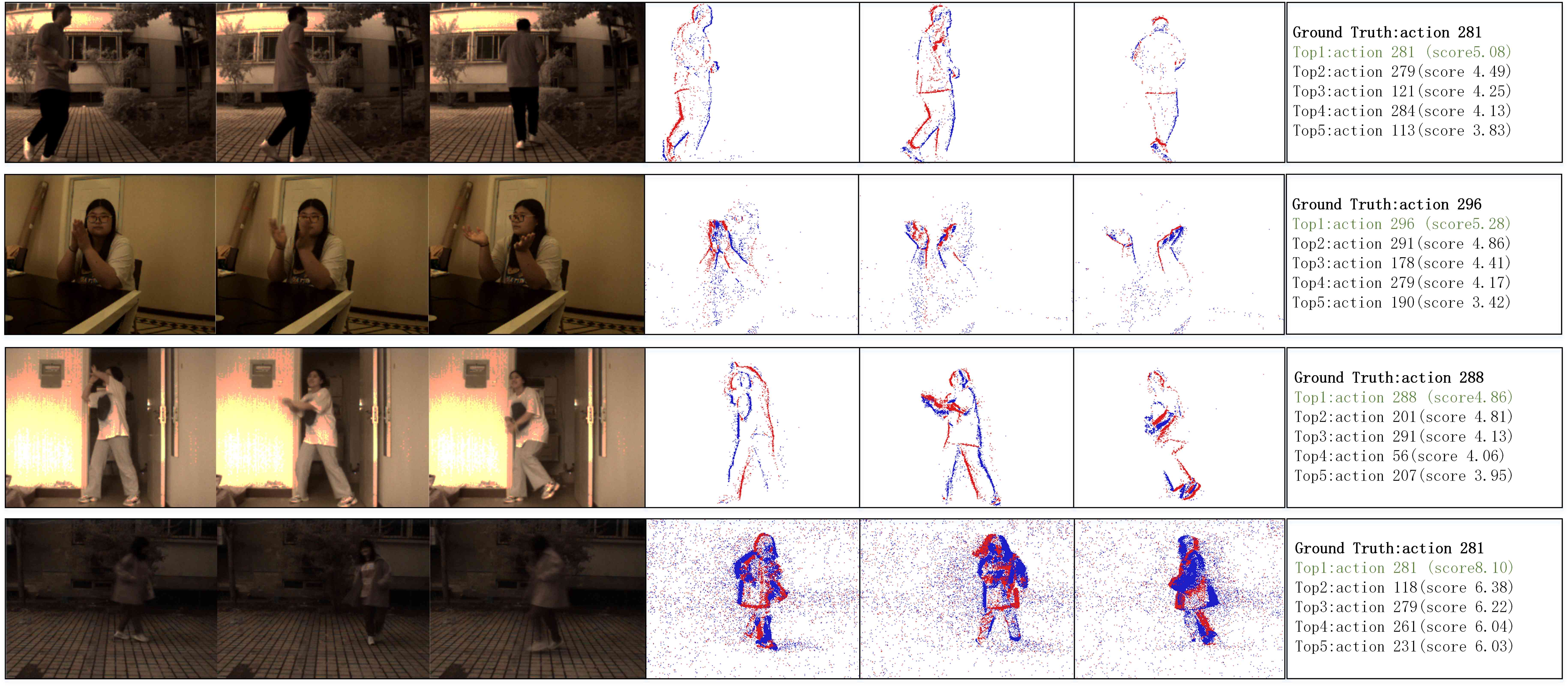}
\caption{Visualization of the top-5 predicted results on the HARDVS dataset.} 
\label{hardvs_top5.jpg}
\end{figure*}

\subsection{Visualization} 
In addition to the aforementioned quantitative analysis, we also give some visualizations of the feature maps, feature distribution, and predicted results. 

\noindent
\textbf{Feature Map. }  
As shown in the first four columns of Fig.~\ref{SCNN_feature_map}, we give a visualization of event streams of our newly proposed PokerEvent dataset and the HARDVS dataset. The following four columns are feature maps extracted from SCNN and DCN modules. We can find that fine-grained texture information is captured well using the \textcolor{black}{SCNN} encoder, while the DCN features focus more on high-level feature representations.

For the features learned in our memory support Transformer network, we give a visualization in Fig.~\ref{RES_MST} to check the distance between different classes. Note that 10 classes are selected for this feature visualization. We can find that our proposed MST module further enhanced the ResNet feature significantly. These visualizations fully validate the effectiveness of our proposed modules to handle the RGB frames and event streams.

\noindent 
\textbf{Recognition Results. } 
We present visualizations of four groups of RGB-event samples and their corresponding top-5 recognition results for two datasets. 
\textcolor{black}{Specifically, Fig.~\ref{pokerevent_top5.jpg} shows samples from the PokerEvent dataset, while the Fig.~\ref{hardvs_top5.jpg} is from the HARDVS dataset.} 
From these figures, we can find that the RGB modality is easily influenced by motion blur, and the event stream captures the motion information well and easily filters out the static background information. Our recognition results also demonstrate that the proposed new framework works well on the RGB-Event based classification problem.

\subsection{Limitation Analysis}
Although our proposed RGB-Event object recognition framework works well on two public benchmark datasets and achieves  a better trade-off between the recognition performance and energy consumption, its overall performance is still limited when compared with other state-of-the-art models. Inspired by the great success of pre-trained multimodal big models~\cite{wang2023mmptms}, we think the recognition performance can be further improved by pre-training our proposed framework on a large-scale RGB-Event dataset in a self-supervised manner. As the RGB frames and event streams are aligned well when output from the DVS346 event camera, the spatial and temporal information between the dual modalities can be used to build the learning objectives. 
On the other hand, our newly proposed method is a hybrid SNN-ANN recognition framework. In future works, we will consider designing pure spiking neural networks to save more energy consumption.

\section{Conclusion}\label{Conclusion} 
In this paper, we propose a novel pattern recognition framework that aggregates the RGB frames and event streams effectively. It contains four main modules, including memory support Transformer (MST), spiking convolutional neural network (\textcolor{black}{SCNN}), multi-modal bottleneck fusion (MBF) module, and prediction head. More in detail, the MST is used to process the RGB frames in a support-query interactive learning manner. The GRU network is also introduced to exploit the temporal information between the support frames. The SCNN is a hybrid SNN-ANN branch that can handle the raw event streams directly. The analytical experimental results show that the used \textcolor{black}{SCNN} encoder significantly saves energy consumption compared with its corresponding ANN version. In addition, we also introduce the multi-modal bottleneck fusion module to combine the dual features more effectively. 

Moreover, due to the dearth of RGB-Event recognition datasets, 
we also propose a new large-scale PokerEvent dataset to bridge the data gap. We conduct extensive experiments on two large-scale datasets to demonstrate that our model can achieve good recognition performance. 
In our future works, we will consider further improving the overall recognition performance based on large-scale self-supervised pre-training. The exploration of pure spiking neural networks will also be an interesting direction.

\small{ 
\bibliographystyle{IEEEtran}
\bibliography{reference}
}

\end{document}